\documentclass[11pt]{article}

\usepackage[final]{acl}

\usepackage{times}
\usepackage{latexsym}
\usepackage{float}
\usepackage{multicol}

\usepackage[T1]{fontenc}
\usepackage[utf8]{inputenc}

\usepackage{microtype}

\usepackage{inconsolata}

\usepackage{graphicx}

\title{Improving Romanian LLM Pretraining Data\\using Diversity and Quality Filtering}

\author{Vlad Negoiță \and Mihai Masala \and Traian Rebedea \\
        National University of Science and Technology POLITEHNICA Bucharest,\\313 Splaiul Independentei, 060042, Bucharest, Romania}

\begin{document}
\maketitle
\begin{abstract}
Large Language Models (LLMs) have recently exploded in popularity, often matching or outperforming human abilities on many tasks. One of the key factors in training LLMs is the availability and curation of high-quality data.
Data quality is especially crucial for under-represented languages, where high-quality corpora are scarce. In this work we study the characteristics and coverage of Romanian pretraining corpora and we examine how they differ from English data. By training a lightweight multitask model on carefully LLM-annotated Romanian texts, we are able to analyze and perform multi-level filtering (e.g., educational value, topic, format) to generate high-quality pretraining datasets. Our experiments show noteworthy trends in the topics present in Romanian and English data, while also proving the effectiveness of filtering data through improved LLM pretraining performance across multiple benchmarks.
\end{abstract}

\section{Introduction}

Recent advances in Artificial Intelligence, especially in Natural Language Processing (NLP), have been driven by Transformer architectures~\citep{NIPS2017_3f5ee243} and Large Language Models (LLMs). These innovations have transformed how machines process language, enabling applications such as conversational AI, intelligent search, machine translation, and content generation.

The success of modern LLMs depends heavily on large, high-quality pretraining datasets~\cite{longpre-etal-2024-pretrainers}. With billions of parameters, these models require vast data to capture statistical patterns, semantic nuances, and world knowledge in human language. Although multilingual datasets~\citep{penedo2024fineweb-2, nguyen2023culturaxcleanedenormousmultilingual, oscar} have driven broad NLP progress, language-specific resources are vital for robust performance across diverse languages. Romanian is under-represented in large multilingual corpora, causing models trained on high-resource languages to underperform on Romanian. Dedicated Romanian datasets are therefore essential for training, finetuning, and evaluating language models~\citep{masala-etal-2024-vorbesti}.

Recent studies highlight data quality as crucial for strong benchmark performance~\citep{lozhkov2024fineweb-edu,ali2025jql,bai2025efficient}. The FineWeb-Edu framework~\citep{lozhkov2024fineweb-edu} proposed an educational-value-based filtering method effective across various tasks, forming the basis for our Romanian adaptation. To understand potential topic bias from this filtering, we also incorporate additional signals to examine diversity issues. Our entire recipe, including code, data, and models is publicly available\footnote{\url{https://huggingface.co/collections/OpenLLM-Ro/pretraining-datasets}}. Our contributions can be summarized as follows:

\begin{itemize}
    \item We build multidimensional (i.e., educational value, topic, format, reader education level) resources: a small human-annotated dataset (100 samples) and a large (1M samples) LLM-annotated dataset. 
    \item We train a lightweight multi-head classifier that enables cost-effective filtering and cross-lingual distribution analysis at scale.
    \item We build the first high-quality pretraining dataset for Romanian - \textbf{FineWeb2-Edu-Ro} - and prove its usefulness by performing continual pretraining. Compared to other approaches, models trained on our filtered dataset exhibit superior performance across a variety of benchmarks.
\end{itemize}

\section{Related Work}

Recent efforts in pretraining focus on high-quality data, combining filtering (rule-based or ML-driven) with growing interest in synthetic data for its significant benefits. FineWeb~\citep{penedo2024the}, FineWeb2~\citep{penedo2024fineweb-2}, and FineWeb-Edu~\citep{lozhkov2024fineweb-edu} represent a collection of high-quality web-based datasets for training large language models. FineWeb initiative offers both English-only datasets (FineWeb - 15T tokens) and multilingual datasets (FineWeb2 - 1000+ languages - 35B Romanian words), cleaned and deduplicated.

\textbf{Quality Filtering.} FineWeb-Edu employs classifier-based quality (educational content) filtering using Llama-3-70B~\citep{llama3modelcard} annotations and Snowflake-Arctic-Embed~\citep{merrick2024arcticembedscalableefficientaccurate} to train a lightweight regressor for educational value.
JQL~\citep{ali2025jql} introduces a language-agnostic setup for annotating educational value in text, using manual labeling and automatic translation to build a multilingual dataset. They benchmark several LLMs and select the top three based on Spearman correlation with human labels, then train lightweight models using Snowflake-Arctic-Embed. Instead of relying on a single metric for data quality,  MetaRater~\citep{zhuang2025metaratermultidimensionaldataselection} uses classifiers for 4 key criteria: professionalism, readability, reasoning, and cleanliness. Reliable identification of lower-quality documents allows alternative approaches such as document rewriting to increase the overall quality of datasets~\citep{recyclingweb}.

% FineWeb~\citep{penedo2024the}, FineWeb2~\citep{penedo2024fineweb-2}, and FineWeb-Edu~\citep{lozhkov2024fineweb-edu} are high-quality web datasets for training large language models. FineWeb is a 15T-token English corpus from CommonCrawl\footnote{\url{https://commoncrawl.org}}, cleaned and deduplicated for LLM performance. FineWeb2 scales this approach to more than 1,000 languages, including 33B Romanian words across 54M documents. FineWeb-Edu focuses on educational content, using Llama-3-70B~\citep{llama3modelcard} annotations and Snowflake-Arctic-Embed~\citep{merrick2024arcticembedscalableefficientaccurate} to train a lightweight regressor for educational value.

% MetaRater~\citep{zhuang2025metaratermultidimensionaldataselection} argues against relying on a single metric for data filtering. They propose building classifiers for 4 key criteria: professionalism, readability, reasoning, and cleanliness. Recycling the Web~\citep{recyclingweb} suggests that aggressive filtering has its limitations: the organic web does not grow as fast as the compute power (high-quality organic data becomes a limitation). They propose using relaxed filtering conditions and the rewriting of the lower-quality documents using Llama-3-70B and chain-of-thought.

\textbf{Romanian Datasets.} Besides FineWeb2, we identify only two other important datasets for Romanian, namely CulturaX~\citep{nguyen2023culturaxcleanedenormousmultilingual} (40B tokens in Romanian) and FuLG~\citep{bădoiu2024fulg150bromaniancorpus} (150B tokens in Romanian). All three datasets stem from CommonCrawl, with different number of snapshots used and different rules for processing and filtering. Crucially, all datasets employ rather standard rules based on n-gram frequency, stop word ratio or text length, and thus lack a more high-level quality based filtering. 

% CulturaX~\citep{nguyen2023culturaxcleanedenormousmultilingual} is a large multilingual dataset (6.3T tokens, 167 languages) using heuristic filters. FuLG~\citep{bădoiu2024fulg150bromaniancorpus}, with 150B carefully curated tokens, is among the largest curated pretraining datasets in Romanian. Both apply heuristic filtering based on stop word ratio, perplexity, text length, and n-gram frequency.

% Several pretraining efforts evaluate performance across different model sizes (1B–7B) using metrics like perplexity and accuracy on benchmarks such as MMLU~\citep{mmlu}, ARC~\citep{arc}, HellaSwag~\citep{hellaswag}, and OpenBookQA~\citep{openbookqa}. FineWeb-Edu trains a 1.71B model on up to 350B tokens and improves over FineWeb, RefinedWeb~\citep{refinedweb}, and C4~\citep{c4} on MMLU, ARC, and OpenBookQA. JQL pretrains a 2B Llama-based model, outperforming FineWeb2 on various benchmarks. FuLG trains a 1B decoder-only OLMo~\citep{olmo} model and reports perplexity gains over mC4 and OSCAR~\citep{oscar}. Recycling the Web uses multiple Llama-2 models and evaluates performance on synthetic and DCLM~\citep{dclm} datasets.

\textbf{Evaluation.} Several pretraining efforts evaluate performance across different model sizes (1B–7B) using metrics like perplexity and accuracy on benchmarks such as MMLU~\citep{mmlu}, ARC~\citep{arc}, HellaSwag~\citep{hellaswag}, or OpenBookQA~\citep{openbookqa}. FineWeb-Edu trains a 1.71B model on up to 350B tokens reporting improvements over unfiltered datasets on MMLU, ARC, and OpenBookQA. Similarly, JQL pretrains a 2B Llama-based model, outperforming FineWeb2 on various benchmarks. FuLG trains a 1B decoder-only OLMo~\citep{olmo} model and reports perplexity gains over mC4 and OSCAR~\citep{oscar}. 

Our approach builds upon recent advancements in data curation for LLMs, but distinguishes itself by jointly predicting multiple signals for a given text in Romanian, as educational value is insufficient for efficient pretraining~\citep{bai2025efficient}.

\section{Taxonomy definition}

For educational quality, we utilize the validated 5-point grading scale from FineWeb-Edu for its effectiveness and to enable comparison between Romanian and English distributions. Regarding additional signals, we follow the taxonomy of WebOrganizer~\citep{wettig2025organizewebconstructingdomains}, which developed topic\footnote{\url{https://huggingface.co/WebOrganizer/TopicClassifier-NoURL}} (e.g., Health, Politics) and format\footnote{\url{https://huggingface.co/WebOrganizer/FormatClassifier-NoURL}} (e.g., News Article, Creative Writing) classifiers to analyze large-scale English web distributions. Finally, we also extract the required educational level per text (preschool through post-graduate studies), which enables progressive learning or curriculum learning for LLMs~\cite{mukherjee2023orca} and could enhance quality prediction. Full taxonomies are provided in \autoref{sec:taxonomy_app}.

\begin{figure*}[!h]
    \includegraphics[width=\linewidth]{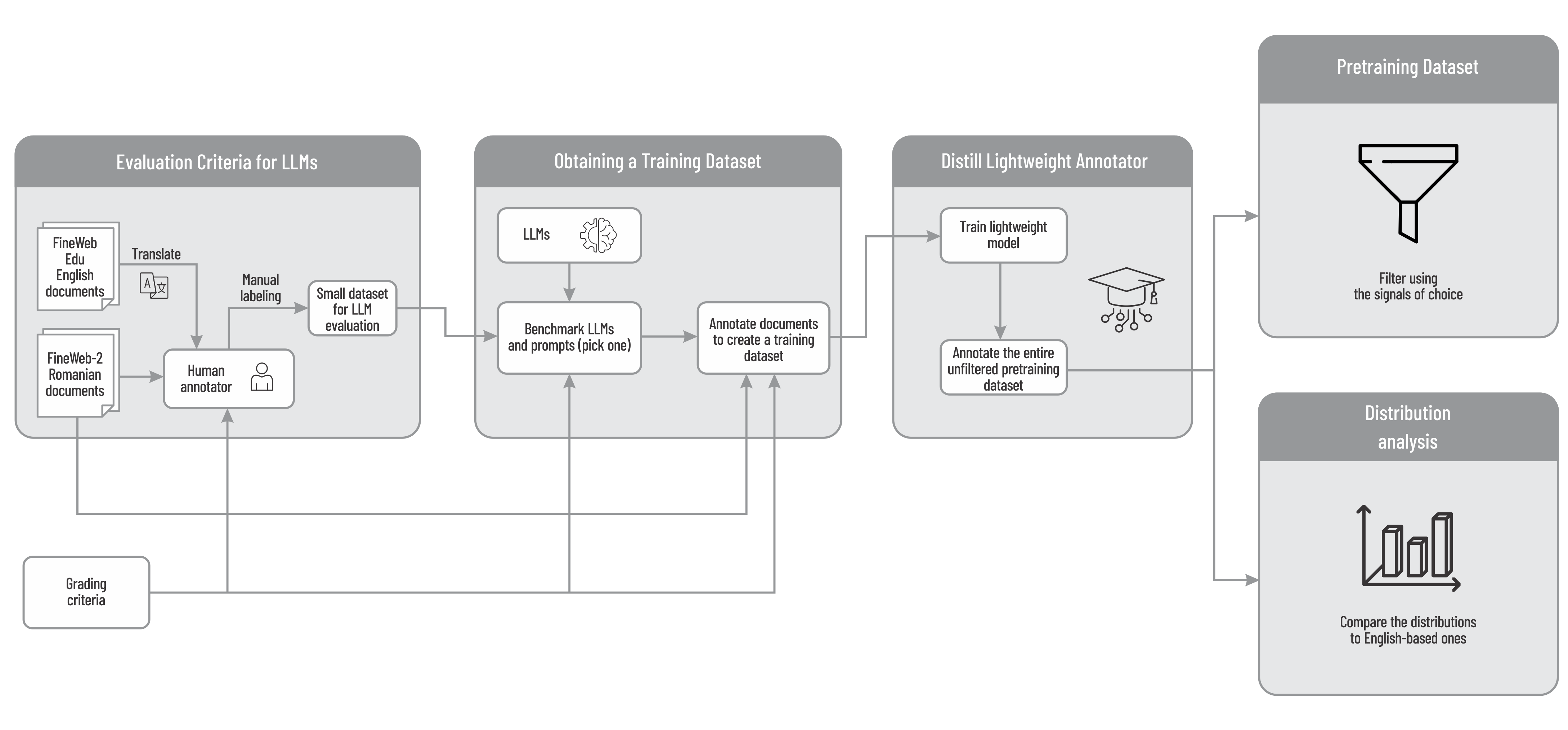}
    \caption{Flowchart detailing the multi-stage pipeline for building an educational Romanian pretraining dataset. The process includes initial human annotation and LLM benchmarking, distillation of a lightweight annotator for full-scale labeling, and final filtering with a subsequent cross-lingual distribution analysis.}
    % \caption{Design and choices overview: choosing texts for human annotation, refining the prompt, choosing the LLM, model distillation, data filtering and distribution analysis.}
    \label{fig:overview}
\end{figure*}

\section{Approach}

% We start by manually annotated a subset of FineWeb2 (containing organic Romanian texts) and FineWeb-Edu (English texts translated into Romanian, retaining their classifier’s educational scores) to enable score alignment and provide a less skewed distribution. This dataset (50 texts from FineWeb2, 50 translated texts from FineWeb-Edu) guided the choice of prompt and LLM for annotating a larger dataset used to distill a lightweight model. \autoref{fig:overview} illustrates the pretraining dataset construction process.

We start by manually annotating a subset of 50 organic Romanian texts from FineWeb2. Furthermore, we add another 50 texts from FineWeb-Edu (translated into Romanian and retaining their classifier’s educational scores) to enable score alignment and provide a less skewed distribution.

We use the resulting dataset (100 samples) to select the best performing LLM (e.g., model, prompt, prompt language) that we will use for annotating a larger dataset, dataset that will be used for training a lightweight classifier. \autoref{fig:overview} illustrates the pretraining dataset construction process.

\subsection{Training dataset}

Following manual annotation, we evaluated a wide range of models, including Llama-3, 3.1, and 3.3, Gemma-2~\citep{gemmateam2024gemma2improvingopen} and Gemma-3~\citep{gemmateam2025gemma3technicalreport}, Cohere-Aya~\citep{aryabumi2024aya}, Qwen-2.5~\citep{qwen2025qwen25technicalreport}, and Mistral-Small~\citep{mistralai2025mistralsmall24b}. We also tested multiple prompting strategies: chain-of-thought, few-shot prompting in both Romanian and English.
When prompted in Romanian with a chain-of-thought approach, Gemma-3 performed exceptionally well for a model of its size (full list of models and results are included in \autoref{sec:llms}). 
% We also found that including a formatting example and placing the educational value right after the explanation significantly improved its performance. 

Thus, we used Gemma-3-12B\footnote{\url{https://huggingface.co/google/gemma-3-12b-it}} to annotate over 1M samples from the Romanian split of FineWeb2.
% \footnote{\url{https://huggingface.co/datasets/HuggingFaceFW/fineweb-2}}.
These examples were partitioned, allocating 10,400 for validation and 20,800 for the test set, with the substantial remainder reserved for the training corpus.

\subsection{Model distillation}

Using the validation split (10,400 examples) and focusing solely on educational value prediction (as a regression task), we selected the encoder architecture, various hyperparameters, and the training set size. We evaluated both Romanian models, such as RoBERT-small, RoBERT-base, RoBERT-large~\citep{masala2020robert} or bert-base-romanian-uncased-v1~\citep{dumitrescu-etal-2020-birth}, and a multilingual model, namely BERT multilingual base~\citep{DBLP:journals/corr/abs-1810-04805}.

We trained a multitask model with four heads: three for classification (topic, format, and educational level) and one for regression (educational value). The model is optimized with a composite loss function, which is a weighted sum of the individual losses from each head. The educational value loss has a weight of $1$, while the three classification losses are weighted by a hyperparameter, $\alpha$. Our experiments showed that this weighting factor did not significantly impact the model's final performance. The final lightweight model configuration is in \autoref{sec:model_setup}, and the results in \autoref{sec:train_corr} confirm that the training data size was sufficient for effective learning.

\section{Results and Discussion}
To create the FineWeb2-Edu-Ro dataset, we augment FineWeb2 with our additional signals, filter for texts with an educational value of 3.5 or higher, and truncate them to a maximum of 4096 tokens. \autoref{tab:thresholds} lists the token and sample counts for the FineWeb2-Edu-Ro and JQL dataset at the chosen filtering thresholds. \autoref{sec:thresholds_impact_sec} presents the impact of various thresholds for our dataset.

\begin{table}[H]
  \centering
  \begin{tabular}{|c|c|c|c|}
    \hline
    \textbf{Data} & \textbf{Threshold} & \textbf{\#Tokens} & \textbf{\#Samples}\\
    \hline
    Ours & 3.5 & 6.43B & 3.9M\\
    JQL  & P92 & 6.23B & 2.7M\\
    \hline
    \end{tabular}
\caption{Comparison of two filtering methods for the FineWeb2 dataset, both truncated at 4096 tokens. Our model used a 3.5 threshold, while JQL's method uses a 92nd percentile threshold on the smallest quantile from its three-headed model's distributions. The thresholds (3.5 and P92) were chosen so that the remaining number of tokens is comparable for the two methods and relatively small (resource constraints).}
\label{tab:thresholds}
\end{table}

\begin{figure}[H]
  \includegraphics[width=\columnwidth]{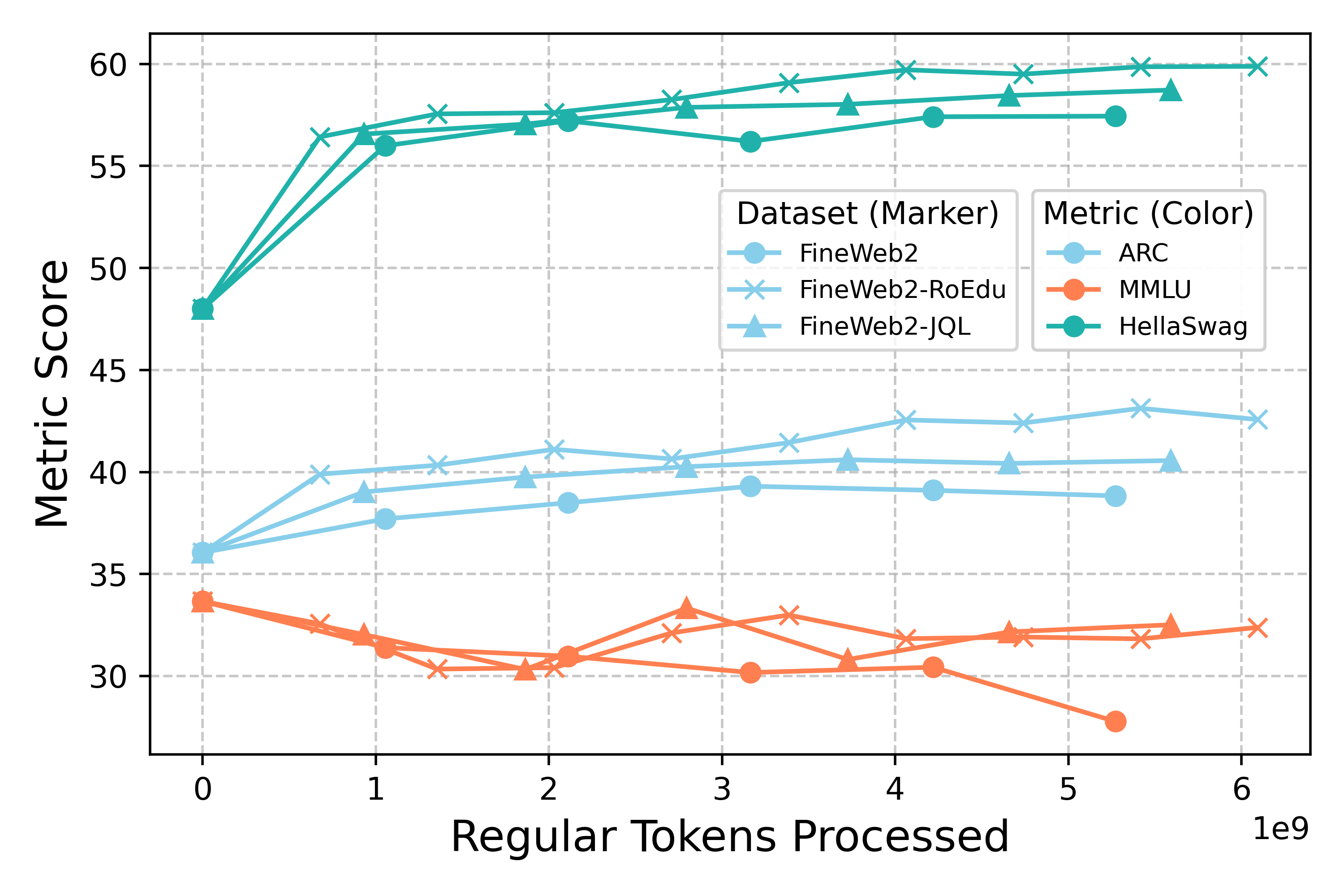}
  \caption{Pretraining results using filtered (RoEdu\&JQL) and unfiltered FineWeb2 data. Note the consistent improvement in performance when using filtered data, with the best overall results obtained using our proposed filtering approach.}
  \label{fig:pretraining}
\end{figure}

We compare the performance of a Llama-2-7B base model~\citep{touvron2023llama2openfoundation} continually pretrained on filtered data versus unfiltered data. Continual pretraining was necessary due to the limited size of the filtered corpus. To ensure fairness, we matched the number of non-padding tokens, accounting for the longer average length of filtered texts. As we propose a method for obtaining higher-quality pretraining datasets for Romanian, we compare our approach with JQL, and we pick FineWeb2 as the underlying dataset to facilitate comparisons with FineWeb-Edu, in terms of topic distribution. Evaluation on Romanian translated~\citep{masala-etal-2024-vorbesti} versions of MMLU~\citep{mmlu}, ARC~\citep{arc}, and HellaSwag~\citep{hellaswag} shows that filtering yields clear performance gains (see \autoref{fig:pretraining}). Both filtering methods improve over the unfiltered FineWeb2 baseline, with our proposed method exhibiting stronger performance on all considered benchmarks. For all metrics, we report the average of multiple few-shot setups: 0, 1, 3, 5 shots for Ro-MMLU, 0, 1, 3, 5, 10, 25 shots for Ro-ARC, and 0, 1, 3, 5, 10 shots for Ro-HellaSwag matching previous work for Romanian~\citep{masala-etal-2024-vorbesti} for a fair comparison. 

\begin{figure}[H]
  \includegraphics[width=\columnwidth]{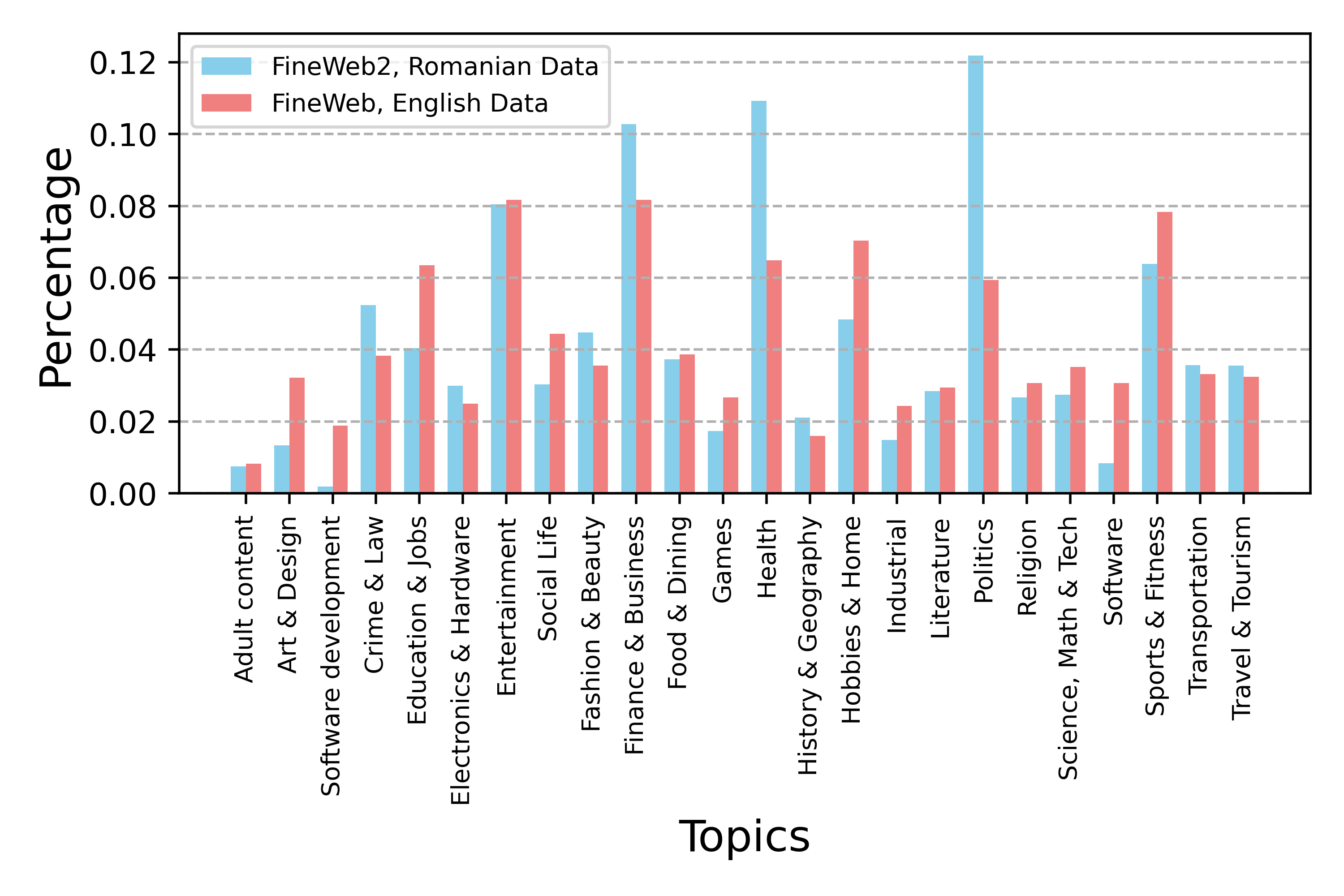}
  \caption{Unfiltered texts topic distribution.}
  % \caption{Topic distributions for unfiltered texts for Romanian and English.}
  \label{fig:fineweb_fineweb2}
\end{figure}

\begin{figure}[H]
  \includegraphics[width=\columnwidth]{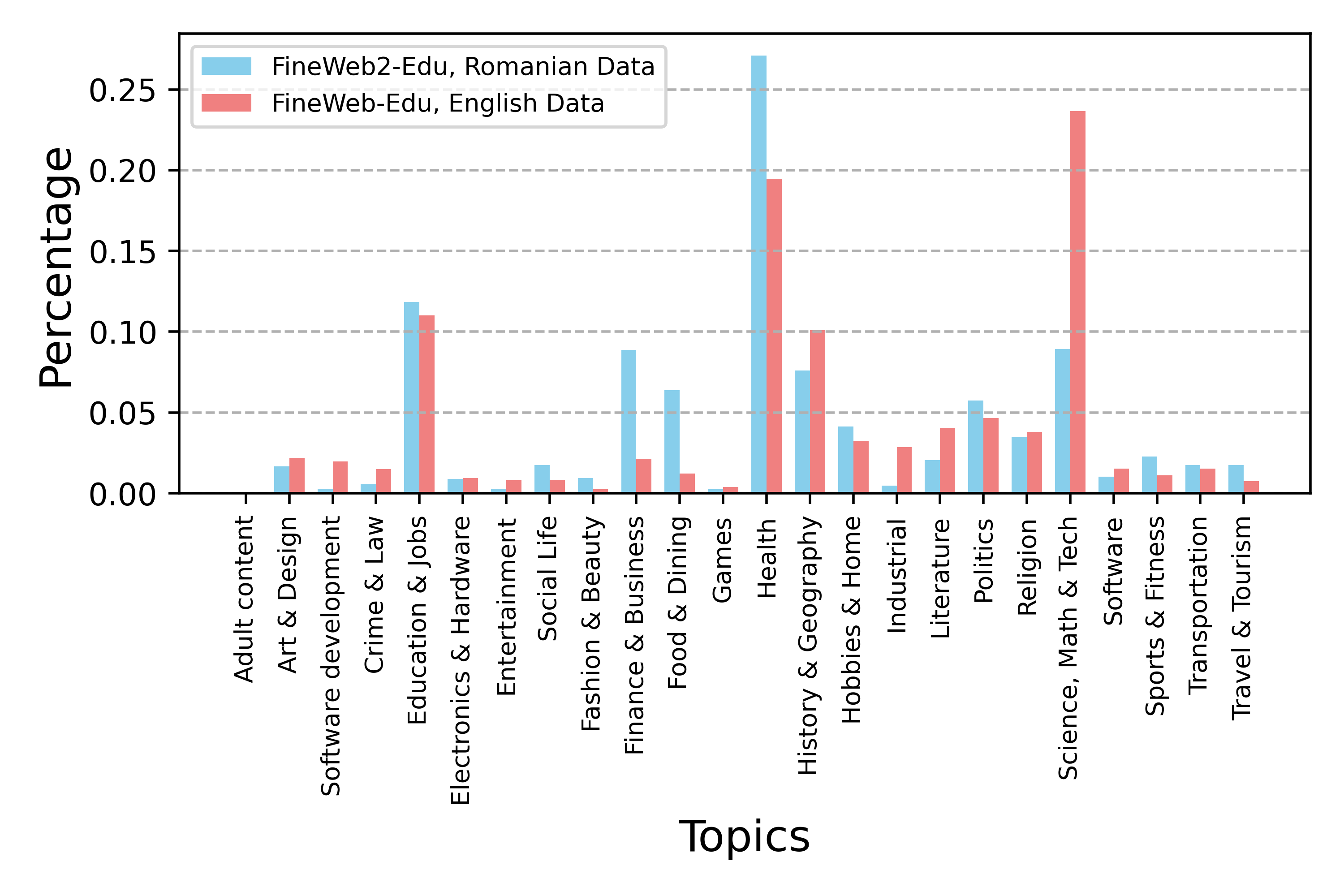}
  % \caption{Topic distributions for filtered texts for Romanian and English.}
  \caption{Filtered texts topic distribution.}
  \label{fig:fineweb_edu_fineweb2_edu}
\end{figure}

\autoref{fig:fineweb_fineweb2} and \autoref{fig:fineweb_edu_fineweb2_edu} show the topic distributions for FineWeb2 (Romanian, annotated with our multitask classifier) and FineWeb (English, annotated using WebOrganizer’s \textit{Topic Classifier no-URL}~\citep{wettig2025organizewebconstructingdomains}) for unfiltered and filtered data (with our method, using a 3.5 threshold). While overall similar, Romanian texts feature higher proportions of \textit{Finance \& Business}, \textit{Health}, and \textit{Politics}, whereas English texts have more \textit{Software}, \textit{Software Development}, and \textit{Education \& Jobs} content.

A comparative analysis following the filtering process reveals a disparity between the datasets. The Romanian corpus is notably deficient in \textit{Science, Math \& Technology} texts compared to its English counterpart, while exhibiting an over-representation of topics such as \textit{Finance \& Business} and \textit{Food \& Dining}. Interestingly, \textit{Food \& Dining} has similar proportions in both unfiltered datasets, but the Romanian classifier amplifies its importance, while the English one diminishes it.

\section{Conclusions}

% We developed a fast, accurate classifier to identify texts with high educational value, by using multilingual LLMs to generate labels and training a lightweight BERT model. Using this classifier to filter datasets such as FineWeb2 significantly improved Llama-2 pretraining, thus proving its effectiveness. We gather signals (topic and format) to both investigate the model's influence on diversity and to facilitate cross-lingual comparisons.

We extend beyond basic educational value extraction by enriching existing large-scale pretraining datasets with detailed metadata on topic, format, and educational level. This approach not only provides finer control over the filtering process but also enables meaningful cross-lingual comparisons. Applying this classifier to datasets such as FineWeb2 has led to substantial improvements in Llama-2 pretraining performance, demonstrating its effectiveness.

\section*{Limitations}

Filtering reduces dataset size below what is needed for optimal large model training per the Chinchilla scaling law~\citep{hoffmann2022trainingcomputeoptimallargelanguage}, so larger datasets are required.

Additionally, limited format diversity in training data affects classifier accuracy. Future work will focus on expanding data diversity.

% add think about risking building on already existing biases

\section*{Acknowledgments}

This research was supported by the project “Romanian Hub for Artificial Intelligence - HRIA”, Smart Growth, Digitization and Financial Instruments Program, 2021-2027, MySMIS no. 351416.

\bibliography{custom}

\appendix

\section{Taxonomy definition}
\label{sec:taxonomy_app}

The comprehensive list of topics and formats belongs to WebOrganizer \cite{wettig2025organizewebconstructingdomains} (we are interested in cross-lingual distribution comparisons, so we use their classifiers for English texts).

Topics (24): \textit{Adult Content}, \textit{Art \& Design}, \textit{Software Development}, \textit{Crime \& Law}, \textit{Education \& Jobs}, \textit{Electronics \& Hardware}, \textit{Entertainment}, \textit{Social Life}, \textit{Fashion \& Beauty}, \textit{Finance \& Business}, \textit{Food \& Dining}, \textit{Games}, \textit{Health}, \textit{History \& Geography}, \textit{Home \& Hobbies}, \textit{Industrial}, \textit{Literature}, \textit{Politics}, \textit{Religion}, \textit{Science, Math \& Technology}, \textit{Software}, \textit{Sports \& Fitness}, \textit{Transportation}, and \textit{Travel \& Tourism}.

Formats (24): \textit{Academic Writing}, \textit{Content Listing}, \textit{Creative Writing}, \textit{Customer Support Page}, \textit{Discussion Forum / Comment Section}, \textit{FAQs}, \textit{Incomplete Content}, \textit{Knowledge Article}, \textit{Legal Notices}, \textit{Listicle}, \textit{News Article}, \textit{Nonfiction Writing}, \textit{Organizational About Page}, \textit{Organizational Announcement}, \textit{Personal About Page}, \textit{Personal Blog}, \textit{Product Page}, \textit{Q\&A Forum}, \textit{Spam / Ads}, \textit{Structured Data}, \textit{Technical Writing}, \textit{Transcript / Interview}, \textit{Tutorial / How-To Guide}, and \textit{User Reviews}.

We propose the following classification for education level (6): \textit{Preschool}, \textit{Primary School}, \textit{Middle School}, \textit{High School}, \textit{Bachelor's Degree}, and \textit{Postgraduate}.

\section{Choosing the LLM}
\label{sec:llms}

\autoref{tab:model_eval_reduced} provides the main metrics that influenced the decision of using Gemma3-12B \cite{gemmateam2025gemma3technicalreport} for the creation of the training dataset that was further distilled.

\begin{table}[H]
    \centering
    \begin{tabular}{|c|c|c|c|}
    \hline
    \textbf{Model} & \textbf{Edu.} & \textbf{Topic} & \textbf{Err.} \\
    \textbf{(prompt lang.)} & \textbf{RMSE} & \textbf{Acc.} & \textbf{} \\
    \hline
    Llama3.3-70B (en) & 1.00 & 0.72 & \textbf{0} \\
    Llama3.3-70B (ro) & 1.48 & 0.72 & \textbf{0} \\
    Llama3-70B (en) & 1.42 & 0.71 & 21 \\
    Llama3-70B (ro) & 1.61 & 0.69 & 35 \\
    Llama3.1-70B (en) & 1.25 & \textbf{0.73} & \textbf{0} \\
    Llama3.1-70B (ro) & 1.37 & 0.70 & 2 \\
    Llama3.1-8B (en) & 1.03 & 0.52 & 5 \\
    Llama3.1-8B (ro) & 1.23 & 0.45 & 17 \\
    Llama3-8B (en) & 1.21 & 0.49 & 8 \\
    Llama3-8B (ro) & 1.33 & 0.56 & 34 \\
    Gemma2-27B (en) & 0.97 & 0.61 & 27 \\
    Gemma2-27B (ro) & 1.04 & 0.58 & 23 \\
    Gemma2-9B (en) & 0.99 & 0.57 & 71 \\
    Gemma2-9B (ro) & 1.06 & 0.60 & 16 \\
    Gemma3-27B (en) & 1.08 & \textbf{0.73} & \textbf{0} \\
    Gemma3-27B (ro) & 1.26 & \textbf{0.73} & 1 \\
    Gemma3-12B (en) & 1.02 & 0.68 & 1 \\
    Gemma3-12B (ro) & \textbf{0.96} & 0.69 & \textbf{0} \\
    CohereAya-35B (en) & 1.26 & 0.60 & 10 \\
    CohereAya-35B (ro) & 1.35 & 0.50 & 29 \\
    Qwen2.5-72B (en) & 1.35 & 0.72 & \textbf{0} \\
    Qwen2.5-72B (ro) & 1.38 & 0.70 & 6 \\
    Mistral-S-24B (en) & 1.07 & \textbf{0.73} & 5 \\
    Mistral-S-24B (ro) & 1.03 & 0.70 & 7 \\
    \hline
    \end{tabular}

\caption{Evaluation of multiple models with both Romanian and English prompts on a manually annotated dataset. Reported metrics include root mean squared error for educational value, accuracy for topic classification, and the number of errors (instances where the model disregarded the instructions).}
\label{tab:model_eval_reduced}
\end{table}

\section{Impact of Training Data over Model Performance}
\label{sec:train_corr}

\begin{figure}[H]
  \includegraphics[width=\columnwidth]{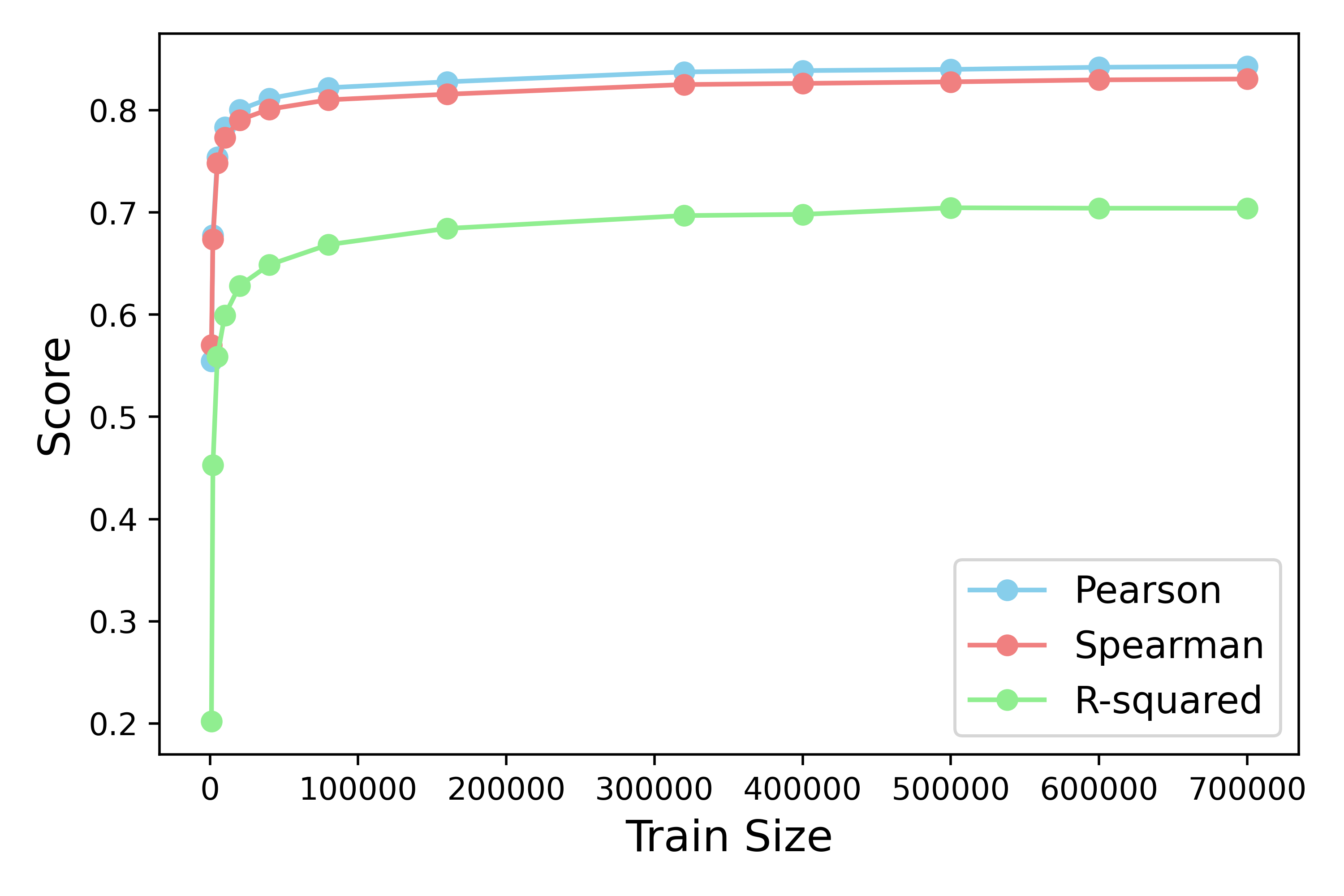}
  \caption{Educational value scores against training size.}
  \label{fig:train_corr}
\end{figure}

\autoref{fig:train_corr} presents how the three considered metrics (Pearson correlation, Spearman correlation, and $R^2$) improve on the validation set as the training dataset size increases.

\section{Lightweight Classifier Setup}
\label{sec:model_setup}

\autoref{tab:model_summary} summarizes the lightweight model architecture.

\begin{table}[H]
  \centering
  \begin{tabular}{|c|c|}
    \hline
    \textbf{Configuration Parameter} & \textbf{Value} \\
    \hline
    Encoder & RoBERT-base \\
    Additional layer size & 256 \\
    Tasks & All \\
    Learning Rate & $1 \cdot 10^{-4}$ \\
    Encoder Learning Rate & $3 \cdot 10^{-6}$ \\
    Training Set Size & 1M \\
    Epochs & 3 \\
    $\alpha$ (loss parameter) & 0.8 \\
    \hline
\end{tabular}
\caption{Lightweight classifier hyperparameters.}
\label{tab:model_summary}
\end{table}

\section{Impact of Filtering Thresholds}
\label{sec:thresholds_impact_sec}

The results in \autoref{tab:thresholds_impact} demonstrate that applying different filters drastically changes the size of the dataset. As expected, there are far fewer high-quality texts available than low-quality ones.

\begin{table}[H]
  \centering
  \begin{tabular}{|c|c|c|}
    \hline
    \textbf{Threshold} & \textbf{\#Tokens} & \textbf{\#Samples}\\
    \hline
    2.0 & 31.60B & 18.7M\\
    2.5 & 22.55B & 12.0M\\
    3.0 & 15.23B & 7.3M\\
    3.5 & 9.15B & 3.9M\\
    4.0 & 2.66B & 1.0M\\
    \hline
    \end{tabular}
\caption{Token counts for multiple filtering thresholds, without any truncation.}
\label{tab:thresholds_impact}
\end{table}

\section{Educational Level Distribution}

\begin{figure}[H]
  \includegraphics[width=\columnwidth]{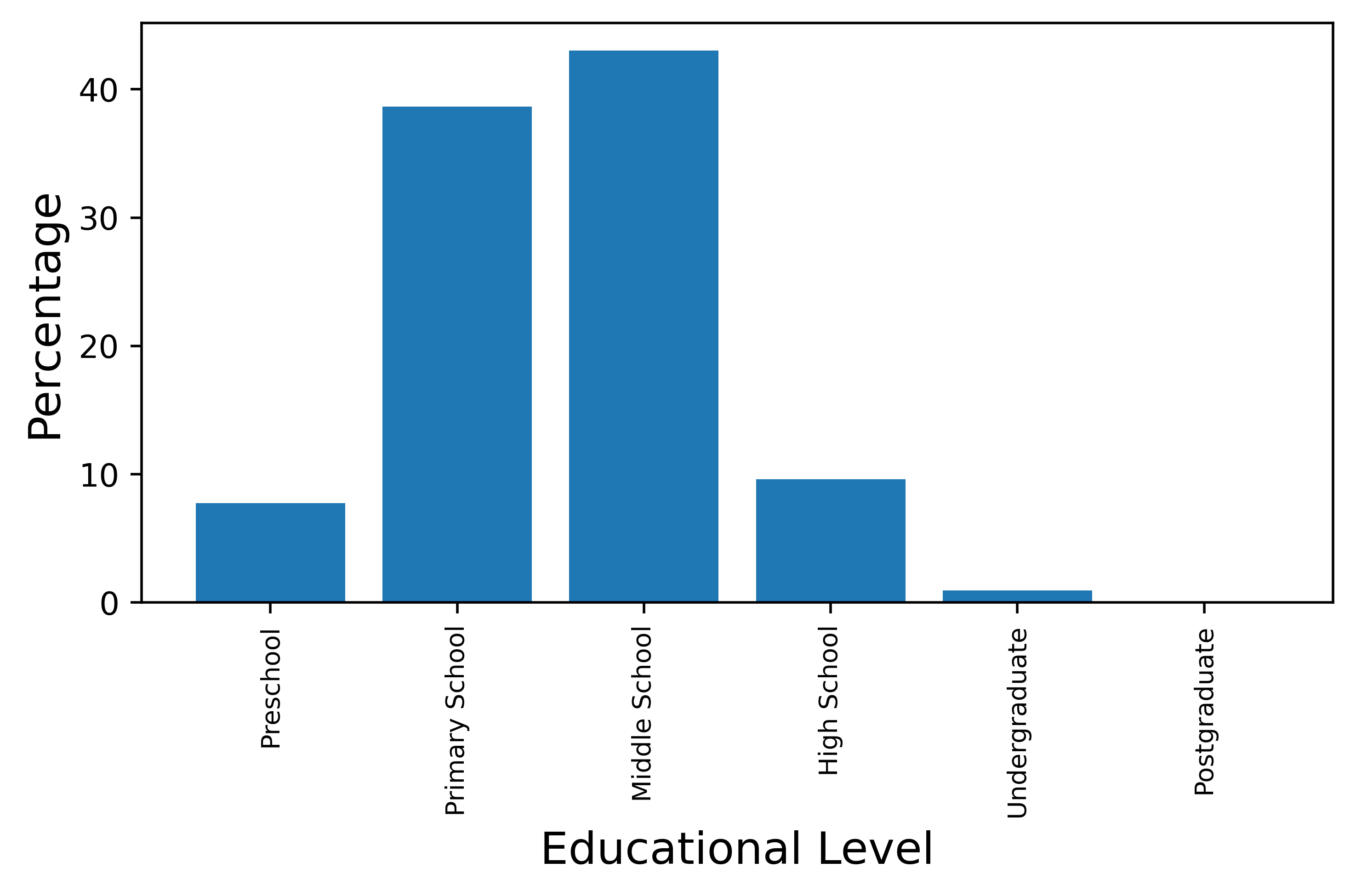}
  \caption{Educational level distribution of FineWeb2 Romanian split.}
  \label{fig:edu_level}
\end{figure}

In \autoref{fig:edu_level}, we present the distribution of educational levels as predicted by our trained classifier on the FineWeb2 Romanian split. A substantial majority of the data (over 80\%) corresponds to primary and middle school levels. In contrast, data corresponding to education levels beyond high school account for only about 1\%. 

\begin{figure}[H]
  \includegraphics[width=\columnwidth]{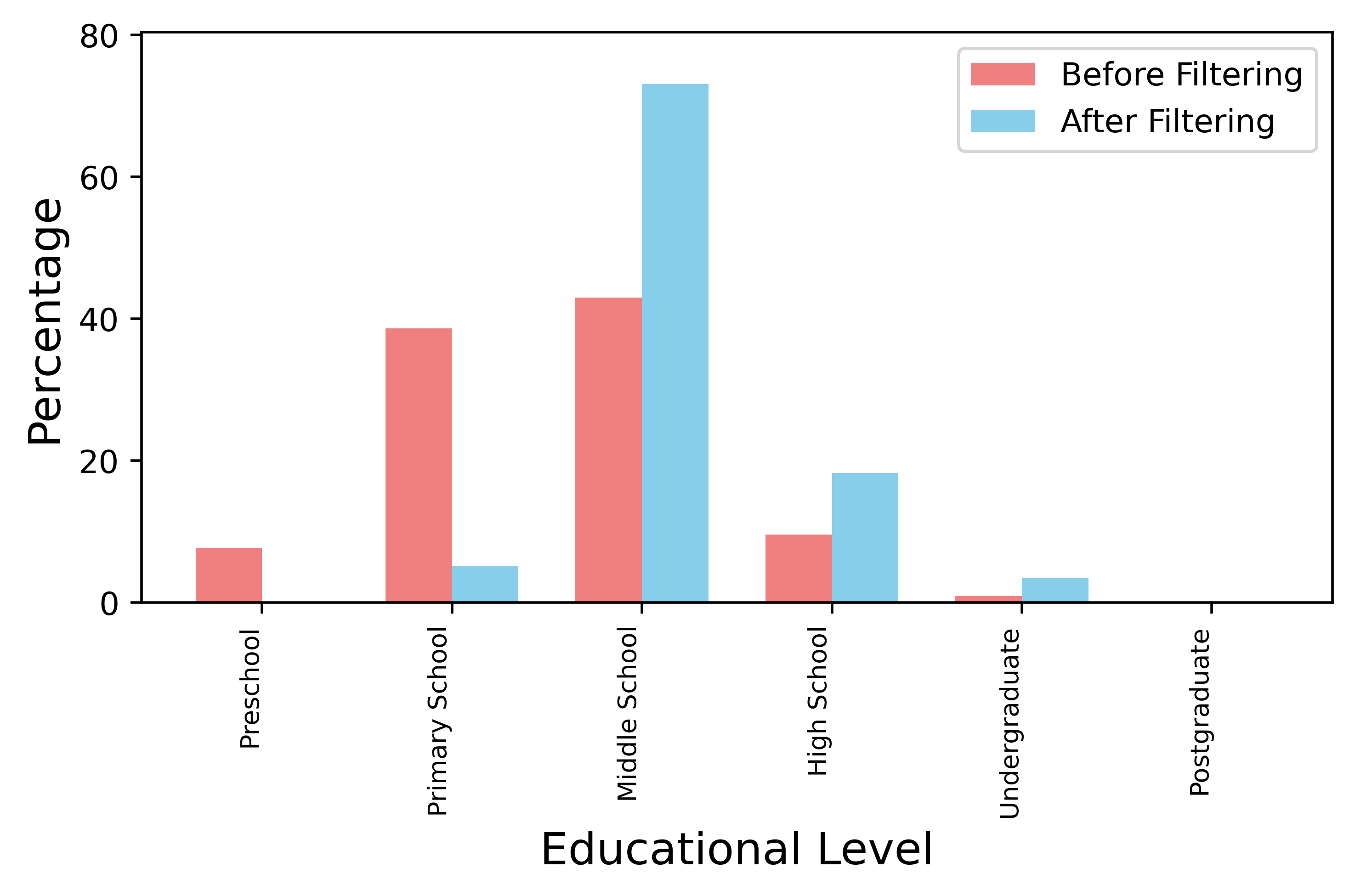}
  \caption{Educational level distribution of FineWeb2 Romanian split before and after filtering.}
  \label{fig:edu_level_after_filtering}
\end{figure}

Distribution of educational level after filtering (with the same 3.5 threshold on educational value) is presented in \autoref{fig:edu_level_after_filtering}. We note that, documents related to lower educational levels (preschool and primary school) are largely filtered out, resulting in a notable decrease in their overall proportion, whereas those associated with higher educational levels are more prevalent in the dataset. 

\section{Training Details}

All training runs were performed on an NVIDIA DGX H100 equipped with 8 GPUs. Continual pretraining on the unfiltered data needed 656 GPU hours, while training on FineWeb2-RoEdu (our filtered data) needed 608 hours with training on FineWeb2-JQL requiring 584 GPU Hours. All training runs were performed using the same hyperparameters: cosine learning rate schedule with $1e-4$ peak and $1e-5$ minimum learning rate and $5\%$ warmup ratio; $4096$ cutoff length, no packing; $64$ batch size with $16$ gradient accumulation steps, leading to an effective batch size of $1024$. Due to resource constraints, we perform and report results on a single run.

% add training data gpus etc + hyperparams
% volunteer student, pays, instructions (same as FineWeb?)

\section{Annotation Details}

\begin{figure}[H]
  \includegraphics[width=\columnwidth]{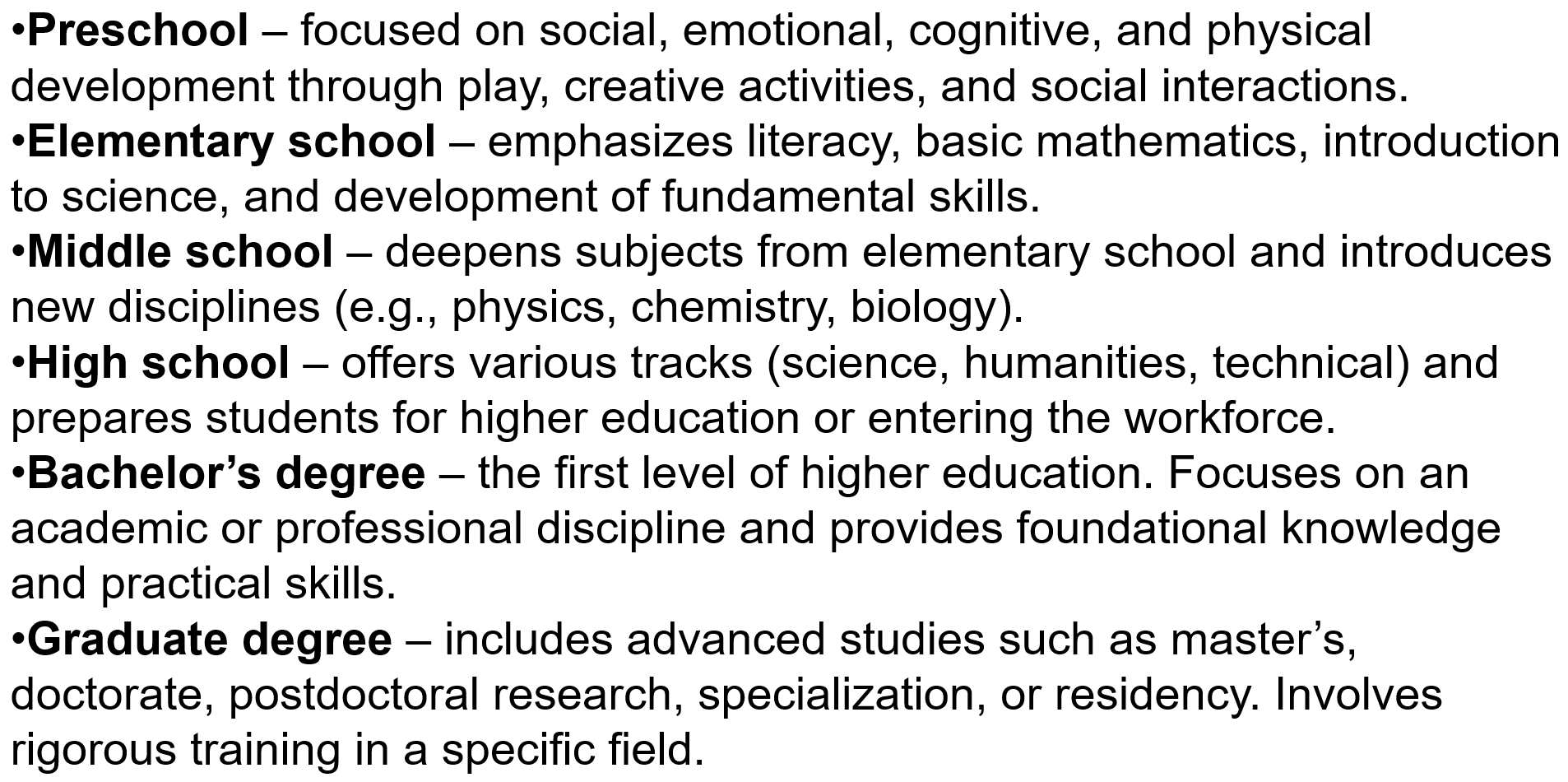}
  \caption{Translated educational level prompt.}
  \label{fig:prompt_edu_level}
\end{figure}

The human annotations (100 documents) were conducted by a Romanian native MSc student in Computer Science who volunteered for the task. For evaluation education value, we translated the FineWeb-Edu English instructions\footnote{\url{https://huggingface.co/HuggingFaceFW/fineweb-edu-classifier/blob/main/utils/prompt.txt}}, while for topic and format we translate the original WebOrganizer prompts~\citep{wettig2025organizewebconstructingdomains}. The educational value prompt, with explanations for each category, as presented (translated) in \autoref{fig:prompt_edu_level}.

\end{document}